\documentclass{bmvc2k}
\usepackage{booktabs}
\usepackage{amssymb}

\title{DOGS: Design-Space Sampling for Prompt-Driven Logo Generation}

\addauthor{Ganyu Zou}{ganyu@vt.edu}{1}
\addauthor{Chen Dai}{chendai@vt.edu}{1}
\addauthor{Nathan Self}{nwself@vt.edu}{1}
\addauthor{Kevin Piper}{kpiper@verisign.com}{2}
\addauthor{Ramachandra Rao Seethiraju}{rseethiraju@verisign.com}{2}
\addauthor{Karthik Shyamsunder}{kshyamsu@verisign.com}{2}
\addauthor{Chang-Tien Lu}{clu@vt.edu}{1}
\addauthor{Naren Ramakrishnan}{naren@cs.vt.edu}{1}

\addinstitution{
 Department of Computer Science\\
 Virginia Tech\\
 Alexandria, Virginia, USA
}
\addinstitution{
 Verisign, Inc.\\
 Reston, Virginia, USA
}
\runninghead{Zou et al.}{DOGS: Design-Space Sampling for Logo Generation}

\def\eg{\emph{e.g}\bmvaOneDot}

\begin{document}

\maketitle

\begin{abstract}
Prompt optimization for text-to-image (T2I) generation has been pursued almost entirely as text rewriting, in which a short user brief is expanded into a longer, model-preferred token sequence. We argue that such a language-space formulation is ill-suited to structured visual design tasks such as logo creation, where a one-line brief leaves most design decisions unspecified. These decisions depend on relational priors that a linear sequence cannot encode, and they leave an uncontrolled channel through which protected marks may be reproduced. We therefore recast logo prompting as sampling within a structured design space, and instantiate this idea as DOGS (Design-space prompting with an Originality-aware GFlowNet Sampler). From a large corpus of real-world logos, we mine a typed, graph-structured design grammar whose edges record empirical co-occurrence. A GFlowNet sampler then generates design graphs with probability proportional to a terminal reward that combines recognizability, aesthetics, and corpus-relative originality. Every slot draws only from a closed design-level vocabulary, and any infringement-inducing or harmful token is removed during parsing. The originality reward further penalizes proximity to existing logos, thereby incorporating infringement avoidance into the method by construction. On two open-source renderers and against nine baselines, DOGS produces logos that are more recognizable and aesthetic, substantially more diverse, and far less prone to trademark infringement.
\end{abstract}
\section{Introduction}
\label{sec:intro}
A text-to-image (T2I) model can render the same idea very differently depending on how the prompt is written. Vague phrasing tends to yield generic or off-target images, whereas a detailed, well-chosen prompt produces far more usable results. Prompt optimization automatically expands a short user input into a more informative prompt, typically by training a language model to rewrite prompts toward human preferences or aesthetic scores \cite{gu_systematic_2023,hao_optimizing_2023,wang_discrete_2024}. Existing work, however, optimizes for general aesthetic appeal, whereas many users want to use these models for professional visual design such as logo creation, a regime in which even the strongest generative models still fall far short of professional standards~\cite{liang_idea-bench_2024}.

In logo design, the users are typically non-experts (e.g. brand owners, startup teams, and independent content creators) whose inputs are necessarily broad descriptions. Yet a usable logo must exhibit strong recognizability, design originality, and conformance to established logo-design conventions, requirements that go beyond generic image quality. A one-line brief thus fixes what the user wants while leaving almost every decision about how to design it unspecified.

Mainstream prompt engineering treats the prompt as free-form text and tunes it toward generic image quality. The methods append modifiers such as style to a brief, but do not decide which entity, layout, or typography the logo should adopt, leaving the essential design choices to the user or to whatever the LLM rewriter implicitly defaults to. We therefore investigate prompt optimization for logo design. Given only a brief user intent, the goal is to automatically generate high-quality prompts that are both well aligned with the preferences of T2I models and faithful to the professional requirements of logo design.
\begin{figure}[t]
\centering
\includegraphics[width=\linewidth]{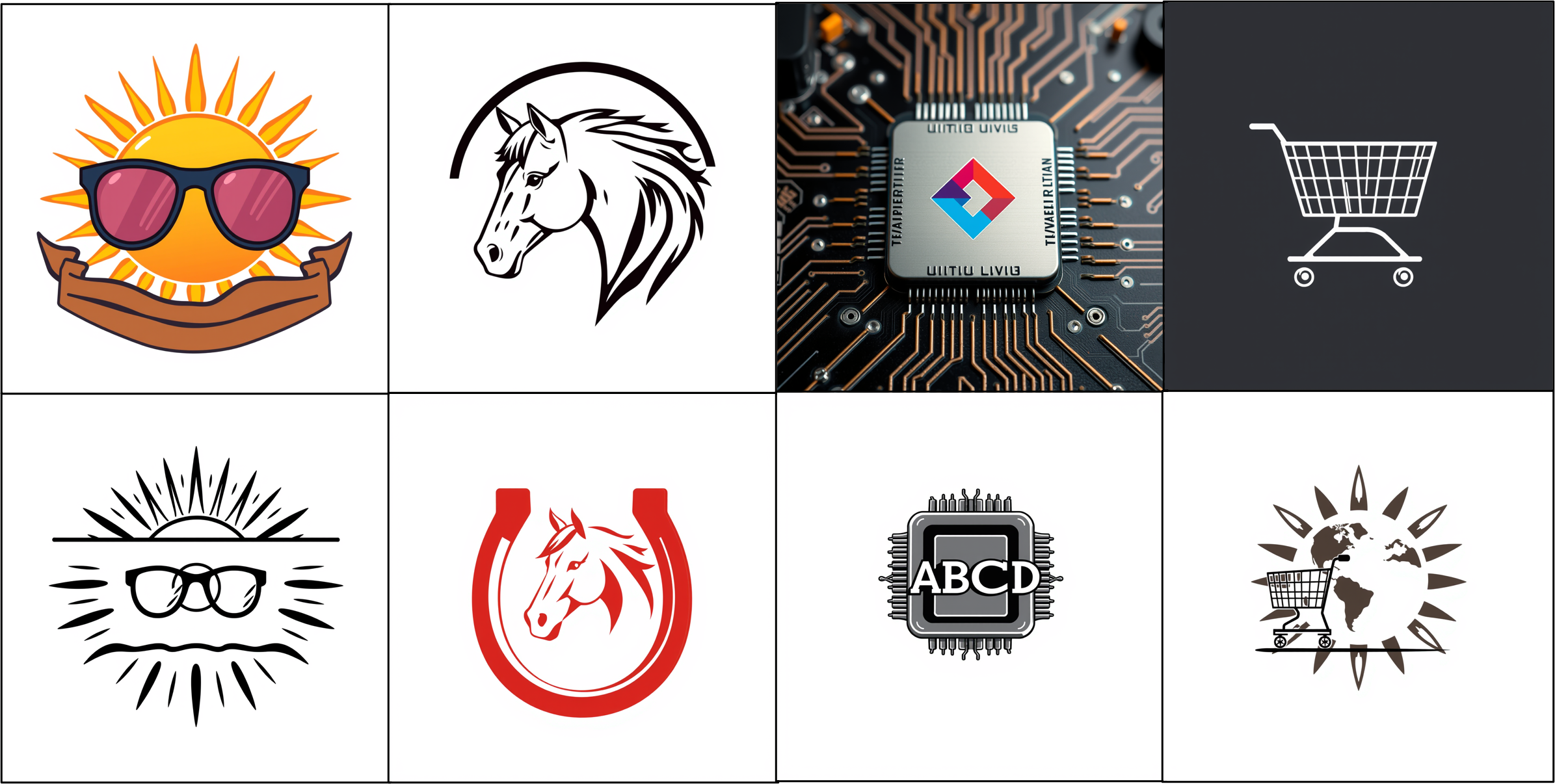}
\caption{\textbf{From user brief to logo design.} Each column is one brief. \emph{Top:} the user prompt rendered directly. \emph{Bottom:} the DOGS output, which fills in the unspecified design decisions to produce a coherent, professionally composed mark.}
\label{fig:teaser}
\end{figure}

This task poses three intrinsic challenges, spanning how a design is represented, how it is searched, and how infringement is controlled. First, \textbf{representing a complete design.} A usable logo requires design decisions that the brief never states, and choosing them professionally rests on relational priors such as which elements should appear together, what role each element should play, and which relations between elements are valid. Such priors cannot be stored in, or recovered from, a linear token sequence. A text-rewriting pipeline therefore has no representation that can hold a complete design, and cannot fill in what the brief leaves unspecified. Second, \textbf{searching the design space.} Logo quality is combinatorial: particular combinations of design elements reliably yield stable, high-quality logos while nearby ones degrade sharply. Moreover, a single brief typically admits many equally valid designs rather than one~\cite{tang_words_2024}. Optimizing a single linear string never exposes the combination itself as the search variable, so existing methods cannot deliberately search for high-quality combinations, and tend to collapse onto a single phrasing rather than cover the many valid designs. Third, \textbf{controlling infringement risk.} Logo generation carries an inherent trademark risk, since an output may reproduce a protected mark. A free-text pipeline provides no explicit objective that measures and penalizes proximity to existing logos, so infringement can only be caught by a post-hoc filter rather than avoided during generation.

{\sloppy To meet these challenges, we propose \textbf{DOGS} (\emph{\textbf{D}esign-space prompting with an \textbf{O}riginality-aware \textbf{G}FlowNet \textbf{S}ampler}), built from three components.} First, we mine a typed, graph-structured design grammar from a large corpus of real-world logos, with nodes as design elements and edges encoding their empirical co-occurrence. This grammar is the representation that linear text lacks, so any sample is a complete design rather than a paraphrase of the brief. Second, the optimization variable becomes a symbolic composition over the grammar, from which the renderer prompt is derived deterministically, and a GFlowNet sampler draws design graphs with probability proportional to a terminal reward, deliberately searching for high-reward combinations while spreading probability mass across the many valid designs rather than collapsing onto one~\cite{bengio_flow_2021,malkin_trajectory_2023}. Third, the design space is a whitelist of purely design-level vocabulary, of which a trademark name is not a member, so such a token has no slot to occupy and is dropped at parsing. The terminal reward then adds a corpus-relative originality term alongside a CLIP recognizability term and an MLLM-judged aesthetics term, penalizing proximity to existing logos so that incidental resemblance is suppressed during sampling rather than caught by a post-hoc filter.

The contributions of this work are summarized as follows:
\begin{itemize}\setlength{\itemsep}{1pt}
    {\sloppy\item \textbf{Design-space prompt optimization:} A reformulation of logo prompt optimization that replaces language-space rewriting with reward-proportional GFlowNet sampling over a typed symbolic design space, exposing the design composition as the search variable and accounting for the many valid designs that a single brief admits.
    \item \textbf{Corpus-induced relational design space:} A typed, graph-structured grammar induced from real-world logos that captures the relational priors (which elements co-occur, in what roles, and under what spatial relations) that are absent from any linear prompt rewriter. This grammar is shared across the action space of the sampler, the corresponding pruning prior, and the originality reward.}
    \item \textbf{Originality-aware reward with infringement-by-construction:} A terminal reward that combines recognizability and aesthetics with a dual-view, corpus-relative originality term spanning pixel-level visual similarity and symbol-level co-occurrence. Together with a design-space whitelist that removes trademark tokens during parsing, this construction embeds infringement avoidance into the sampler rather than relying on post-hoc filters.
    \item \textbf{Evaluation protocol and cross-renderer study:} A logo-specific evaluation protocol that comprises a quality battery and a three-tier infringement battery, which probes brand, company, and design prompts at decreasing levels of explicitness. Using this protocol in a controlled comparison on two open-source renderers against nine baselines, we find that DOGS produces logos that are more recognizable and aesthetic, substantially more diverse, and far less prone to trademark infringement.
\end{itemize}

\section{Related Work}
\label{sec:related-work}

Our work relates to three lines of research: prompt optimization for T2I generation, which addresses the same task in the natural-language space; generation with explicit structure, which makes structural priors a first-class element of the generative pipeline; and reward-driven training of generative models, which tunes the model itself against scalar rewards rather than refining its input.

\paragraph{Prompt Optimization for T2I Generation.}
\label{sec:rw-prompt-opt}
Existing methods operate within the natural-language token space and fall into two families. One rewrites short user inputs into model-preferred prompts with a language model, trained through supervised fine-tuning, reinforcement learning, or preference optimization against aesthetic and human-preference signals~\cite{cao_beautifulprompt_2023,hao_optimizing_2023,mo_dynamic_2024,rosenman_neuroprompts_2024}, with recent variants either decoupling the rewriter from the renderer via reward-agnostic feedback~\cite{kim_reward-agnostic_2025} or chaining the two through co-trained alignment~\cite{wang_promptenhancer_2025}. The other optimizes the prompt as a discrete object directly: through a gradient-based hard-prompt search~\cite{wen_hard_2023}, a human-in-the-loop genetic algorithm over keyword combinations~\cite{pavlichenko_best_2023}, or a ``shortcut text gradient'' propagated through the diffusion denoiser~\cite{wang_discrete_2024}. A broader overview of vision-language prompting is given by~\cite{gu_systematic_2023}.

\paragraph{Generation with Explicit Structure.}
\label{sec:rw-structure}
A second body of work treats explicit structure as a first-class element of generation, differing by where the structure enters the pipeline. One line makes the structured object the generation target itself: sg2im~\cite{johnson_image_2018} synthesizes images from a scene graph, abstract syntax networks~\cite{rabinovich_abstract_2017} and TRANX~\cite{yin_tranx_2018} generate syntax trees for code and semantic parsing, and DiGress~\cite{vignac_digress_2023} samples graphs by discrete denoising diffusion. A second keeps a pretrained decoder intact but restricts its output space at inference to respect a formal grammar, from PICARD for text-to-SQL~\cite{scholak_picard_2021} and grammar-constrained decoding~\cite{geng_grammar-constrained_2023} to later work aligning the constrained distribution with model likelihood~\cite{park_grammar-aligned_2025} or improving its efficiency~\cite{park_flexible_2025}. Closest to ours, a long line in graphics defines a visual design space through a grammar and searches within it: shape grammars~\cite{stiny_introduction_1980} give an early hand-authored formalization, while GRAINS~\cite{li_grains_2019}, LayoutTransformer~\cite{gupta_layouttransformer_2021}, and LayoutDM~\cite{inoue_layoutdm_2023} learn the structure from layout and scene corpora.

\paragraph{Reward-Driven Training of Generative Models.}
A third body of work fine-tunes generative models against scalar rewards rather than refining their input, differing by training objective. One line maximizes an expected reward by reinforcement learning: PPO~\cite{schulman_proximal_2017} gives the canonical clipped objective and RLHF~\cite{christiano_deep_2023,ouyang_training_2022} uses it for preference alignment, adopted by Promptist~\cite{hao_optimizing_2023} and BeautifulPrompt~\cite{cao_beautifulprompt_2023} to fine-tune prompt rewriters. A second removes the explicit reward model and trains directly on preference pairs, via Direct Preference Optimization~\cite{rafailov_direct_2024} and subsequent prompt-optimization variants~\cite{wang_promptenhancer_2025}. A third spreads probability mass across multiple high-reward modes: GFlowNet~\cite{bengio_flow_2021} samples trajectories in proportion to their terminal reward with trajectory balance~\cite{malkin_trajectory_2023} as the canonical objective, PAG~\cite{yun_learning_2025} applies this to T2I prompt optimization over tokens, and related mechanisms pursue diversity through maximum-entropy RL~\cite{haarnoja_soft_2018}, quality-diversity search~\cite{mouret_illuminating_2015}, and novelty search~\cite{lehman_abandoning_2011}. 
\section{Methodology}
\label{sec:method}

\subsection{Overview}
\label{sec:method:overview}
We frame logo prompt engineering as a \emph{design completion} task: given a brief user intent, the system completes the design by sampling a structured symbolic composition that constitutes a full set of design decisions, from which a renderer prompt is derived deterministically and passed to a frozen T2I model.

\paragraph{Task.} Let $\mathcal{S}$ denote the space of symbolic design states (Sec.~\ref{sec:method:space}), $\mathcal{T}$ the space of renderer prompts, and $\mathcal{I}$ the space of rendered images. A deterministic template $\Theta: \mathcal{S} \to \mathcal{T}$ converts a symbolic state into a renderer prompt, and a frozen T2I model $\mathcal{G}: \mathcal{T} \to \mathcal{I}$ renders it into an image. Given a user brief $b$, a parser produces a \emph{partial state} $s_0 \in \mathcal{S}$ that fills whatever design decisions $b$ explicitly specifies and leaves the remaining slots empty. The \emph{design completion} task is to sample a terminal state $s_T \in \mathcal{S}$, consistent with $s_0$, that scores highly under a domain-specific reward $R: \mathcal{S} \to \mathbb{R}_+$ combining a symbolic component for corpus-relative originality on the design state and an image-side component for recognizability and aesthetics on the rendered image $\mathcal{G}(\Theta(s_T))$. We train a sampler $\pi$ whose distribution over terminal states is \emph{proportional} to the reward, $\pi(s_T \mid s_0) \propto R(s_T)$, so as to discover a diverse set of high-reward designs rather than collapsing to a single mode.

\begin{figure}[t]
\centering
\includegraphics[width=0.85\linewidth]{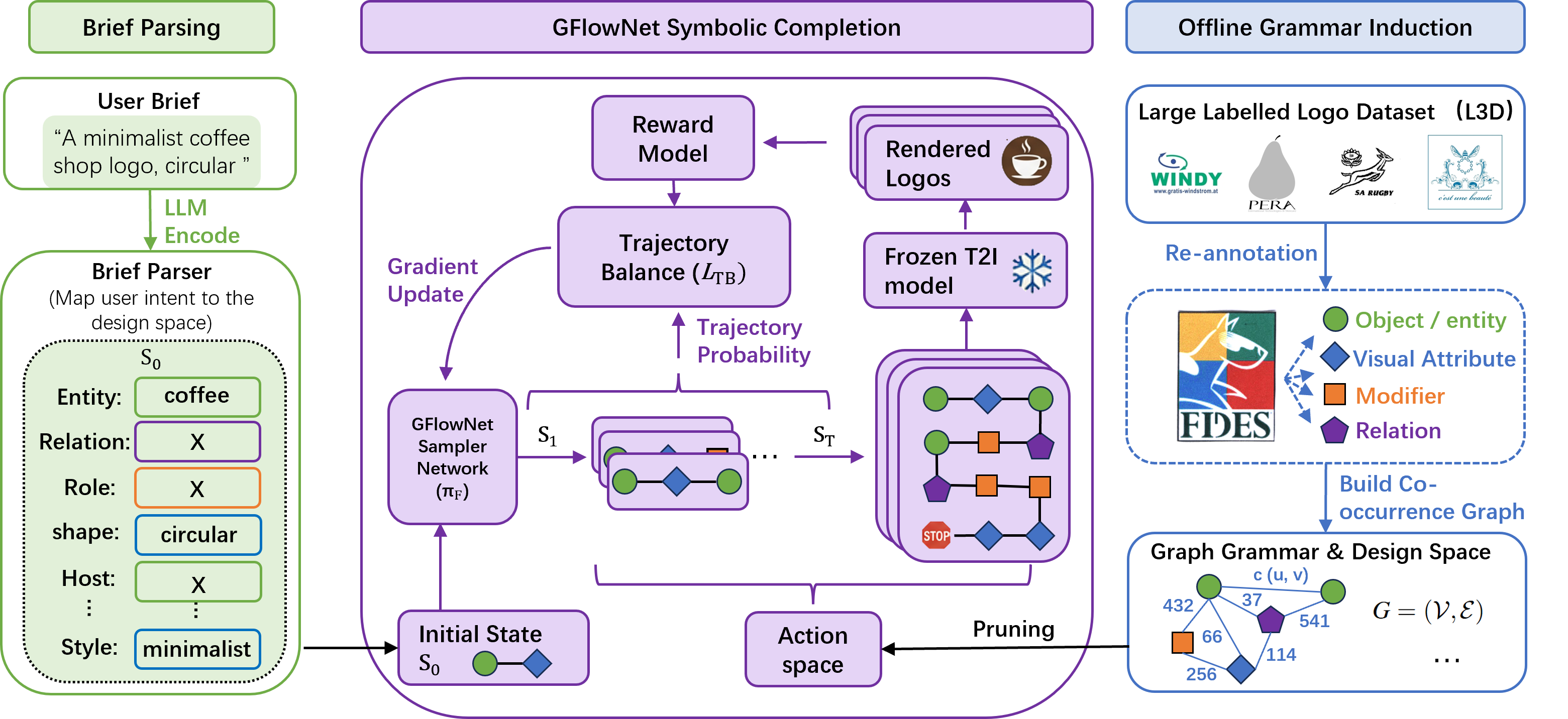}
\caption{\textbf{DOGS pipeline}, organized into three stages. \emph{Brief Parsing} maps the user brief into an initial state $s_0$, filling the design slots it specifies and leaving the rest undecided. \emph{Offline Grammar Induction} re-annotates a logo corpus into typed nodes and builds a co-occurrence grammar $G=(\mathcal{V},\mathcal{E})$, pruned to define the per-step action space. \emph{GFlowNet Symbolic Completion} draws a trajectory $s_0\to\cdots\to s_T$ with the sampler $\pi_F$ over that action space; the terminal state $s_T$ is rendered into a logo by a frozen T2I model, scored by the reward model, and the resulting trajectory probability trains $\pi_F$ via the Trajectory Balance loss $\mathcal{L}_{\text{TB}}$.}
\label{fig:pipeline}
\end{figure}

\paragraph{Pipeline.} Figure~\ref{fig:pipeline} shows the four stages: offline grammar induction from a corpus $\mathcal{D}$ of re-annotated logos, parsing of the brief into $s_0$, symbolic completion by the GFlowNet sampler, and rendering of the terminal state through $\Theta$ and $\mathcal{G}$. The grammar $G$, the corpus statistics behind the originality reward, the renderer $\mathcal{G}$, the CLIP encoders, and the MLLM judge are all fixed. Only the action scorer and a $\log Z(s_0)$ head are trained, isolating prompt-side optimization from pixel synthesis.

\subsection{Symbolic Design Space}
\label{sec:method:space}
We define the symbolic design space $\mathcal{S}$ as the set of structured compositions over a fixed taxonomy of \emph{typed design nodes}. Each node is assigned a type drawn from $11$ categories that together cover the design decisions required by a logo:
\begin{itemize}\setlength{\itemsep}{1pt}
\item Three \emph{object types} for the foreground content: \texttt{entity\_atom} (the subject of the logo), \texttt{host\_atom} (the bounding shape or carrier), and \texttt{letter\_object} (any wordmark or monogram);
\item Five \emph{visual-attribute types}: \texttt{style}, \texttt{layout}, \texttt{shape}, \texttt{typography}, and \texttt{color};
\item Two \emph{element-modifier types}: \texttt{role} (the prominence of an element, taken from \textsf{primary}, \textsf{secondary}, \textsf{support}, \textsf{ornament}, and \textsf{background}) and \texttt{repetition} (single versus repeated instances);
\item One \emph{relation type}: \texttt{relation}, whose value is drawn from a curated vocabulary $\mathcal{L}$ of $23$ spatial and structural labels (\eg, \textsf{above}, \textsf{inside}, \textsf{flanking}, \textsf{attached\_to}, \textsf{around}).
\end{itemize}

\paragraph{State.} A state $s \in \mathcal{S}$ assembles nodes of these types into a structured composition:
\begin{equation}
s = (\mathbf{a},\, \mathbf{o},\, \mathbf{v},\, E),
\end{equation}
where $\mathbf{a} = (a_1, \dots, a_n)$ is a variable-length tuple of \texttt{entity\_atom} nodes, whose per-state cardinality is bounded by a small constant in practice. The tuple $\mathbf{o} = (o_{\text{host}}, o_{\text{letter}})$ holds two fixed slots corresponding to the object types \texttt{host\_atom} and \texttt{letter\_object}. And $\mathbf{v} = (v_{\text{style}}, v_{\text{layout}}, v_{\text{shape}}, v_{\text{typography}}, v_{\text{color}})$ is a tuple of five fixed visual-attribute slots. Let $\mathcal{N}(s) = \{a_1, \dots, a_n\} \cup \{o_{\text{host}}, o_{\text{letter}}\}$ denote the set of object-typed nodes in $s$. Then $E$ is a set of relation triples $(h, \ell, t)$, in which $h$ and $t$ are references to elements of $\mathcal{N}(s)$, and $\ell$ is a node of type \texttt{relation} taking values in $\mathcal{L}$. Each $a_i$ and each $o_{\star}$ is further linked to a single \texttt{role} node and a single \texttt{repetition} node. Any slot, entity, modifier link, or relation triple may be left \emph{empty}, indicating an undecided design choice that the sampler may fill at a later step.

\subsection{Corpus-Induced Design Grammar}
\label{sec:method:grammar}
From the re-annotated corpus $\mathcal{D}$ of Sec.~\ref{sec:method:space}, we mine a fixed graph-structured design grammar $G = (\mathcal{V}, \mathcal{E})$ over the typed nodes of the design space. The grammar both bounds the space of legal next actions at each step and supplies the empirical priors that guide them.

\paragraph{Co-occurrence graph.} The node set $\mathcal{V}$ is the union, across logos $d \in \mathcal{D}$, of every typed node that survives the schema's filtering rules: the five visual-attribute nodes filled by the annotator, the recognised \texttt{entity\_atom}, \texttt{host\_atom}, and \texttt{letter\_object} nodes, the \texttt{role} and \texttt{repetition} nodes attached to those elements, and the canonicalised \texttt{relation} nodes drawn from each logo's annotated relations. We treat each logo $d$ as a set $B_d \subseteq \mathcal{V}$ of the typed nodes that appear in its re-annotated state (each node is counted at most once per logo), and write
\begin{equation}
c(u, v) = \big|\{d \in \mathcal{D} : u, v \in B_d\}\big|,\qquad (u, v) \in \mathcal{V}^{2},
\end{equation}
for the number of logos in which $u$ and $v$ co-occur. The edge set is then
\begin{equation}
\mathcal{E} = \big\{(u, v) \in \mathcal{V}^{2} : c(u, v) \ge \tau\big\},
\end{equation}
where $\tau$ is a small support threshold that filters out low-support co-occurrences; each edge carries the weight $c(u, v)$. The resulting $G$ summarises, type-agnostically, which design elements are empirically likely to appear together in real logos.

\paragraph{Action grammar.} From any partial state $s$, we generate the legal next actions by an enumeration procedure that respects the schema: an action either fills an empty entry in $s$ or terminates the trajectory. We use seven action types, organised into seven groups:
\begin{itemize}\setlength{\itemsep}{1pt}
\item \textsc{add\_entity\_atom\_member}: append a new \texttt{entity\_atom} node to $\mathbf{a}$;
\item \textsc{add\_object\_to\_slot}: fill an empty $o_{\text{host}}$ or $o_{\text{letter}}$ slot with a node of the corresponding object type;
\item \textsc{add\_global\_visual}: fill one of the five empty visual-attribute slots in $\mathbf{v}$;
{\sloppy \item \textsc{add\_role}, \textsc{add\_repetition} (the \emph{modifier} group): attach a \texttt{role} or \texttt{repetition} node to an entity atom or object slot whose corresponding modifier link is empty;
\item \textsc{add\_relation\_triple}: emit a new triple $(h, \ell, t)$ between two distinct occupied refs in $\mathcal{N}(s)$ with a \texttt{relation} node $\ell \in \mathcal{L}$;}
\item \textsc{stop}: terminate the trajectory, yielding $s_{T} = s$.
\end{itemize}

\paragraph{Pruning the per-step candidate set.} After the grammar-legality filter, the candidate set at each step is large (thousands of vocabulary nodes for the entity types alone). Leaving the action space at this scale would force the sampler to spread its mass over a long tail of empirically implausible candidates which would slow credit assignment under the trajectory-balance objective. We prune it with a single score that combines empirical co-occurrence with a redundancy penalty for the entity branch. Let $\text{active}(s) \subset \mathcal{V}$ denote the set of nodes already present in $s$. For a candidate node $u$ to be added at $s$, we define a normalised relevance
\begin{equation}
\rho(u, s) = \frac{1}{m}\sum_{v \in \mathcal{K}_m(u, s)} \frac{\log\!\big(1 + c(u, v)\big)}{\log\!\big(1 + c_{\max}\big)},
\end{equation}
where $\mathcal{K}_m(u, s) \subseteq \text{active}(s)$ denotes the $m$ nodes in $\text{active}(s)$ with the largest co-occurrence count $c(u, v)$ with the candidate $u$, $c_{\max} = \max_{u', v'} c(u', v')$ is the largest pair count in $G$, and we set $m = 2$ in practice. For \textsc{add\_entity\_atom\_member} actions only, we additionally compute a redundancy term against the entity atoms already present in $\mathbf{a}$,
\begin{equation}
\delta(u, \mathbf{a}) = \max_{a_i \in \mathbf{a}}\, \cos\!\big(\phi(u), \phi(a_i)\big),
\end{equation}
where $\phi(\cdot)$ is a frozen CLIP ViT-L/14 text embedding~\cite{radford_learning_2021} of the node's surface text, with a lexical Jaccard-similarity fallback when CLIP features are unavailable. The combined score is
\begin{equation}
\sigma(u, s) = \rho(u, s) - \lambda\, \delta(u, \mathbf{a}),\qquad \lambda = 0.5,
\end{equation}
so that entity-atom proposals follow a Maximal Marginal Relevance (MMR) preference for high co-occurrence with the current state and low similarity to already-chosen entities. For the remaining action groups the redundancy term is omitted and $\sigma(u, s) = \rho(u, s)$. We keep the top-$k$ candidates per action group, and the \textsc{stop} action is always retained. 

\subsection{GFlowNet for Design Completion}
\label{sec:method:sampler}
We cast logo design as a combinatorial optimisation problem in which a usable logo is composed from typed design elements. GFlowNets~\citep{bengio_flow_2021} are well-suited to this formulation: trained to sample compositional objects in proportion to a reward, they cover multiple high-reward modes rather than collapsing to one, matching our requirement of diverse, novel designs.

The sampler is trained in two stages. A supervised \emph{completion pretrain} first learns a forward policy that imitates terminal-consistent action choices on the re-annotated corpus, providing a strong warm start. This policy is then finetuned with a GFlowNet objective such that, at convergence, the sampler draws design states in proportion to the terminal reward defined in Sec.~\ref{sec:method:reward}. Throughout both stages, the policy operates over the pruned action sets $\mathcal{A}(s)$ introduced in Sec.~\ref{sec:method:grammar}.

\paragraph{Supervised completion pretrain.} The pretrain phase uses the re-annotated corpus as direct supervision. For each annotated terminal state $s_T$, a masked intermediate state $s$ is obtained by removing a random subset of its entries. The actions that bring $s$ back toward $s_T$ are labelled as \emph{positive} actions $\mathcal{A}^+(s)$, and the remaining elements of $\mathcal{A}(s)$ as \emph{negatives} $\mathcal{A}^-(s)$. The pretrain loss is a per-action-type contrastive log-likelihood:
\begin{equation}
\mathcal{L}_{\text{pre}}(s) = -\frac{1}{|\mathcal{K}(s)|}\sum_{\kappa \in \mathcal{K}(s)} \log \frac{\sum_{a \in \mathcal{A}^+_{\kappa}(s)} e^{f_\theta(s, a)}}{\sum_{a \in \mathcal{A}^+_{\kappa}(s) \cup \mathcal{A}^-_{\kappa}(s)} e^{f_\theta(s, a)}},
\end{equation}
where $f_\theta(s, a)$ is the scalar logit produced by the policy network for action $a$ at state $s$, and $\mathcal{K}(s)$ is the set of action types $\kappa$ that contain at least one positive and at least one negative action at $s$. The per-state loss is averaged over a batch, yielding a forward policy $\pi_{\text{pre}}$ that captures, type by type, which actions are typical at intermediate states under the corpus distribution. This $\pi_{\text{pre}}$ is then used both to initialise the finetune policy and as a frozen reference for the KL anchor introduced below.

\paragraph{Trajectory-balance objective.} Given $s_0$, the forward policy $\pi_F$ samples a trajectory $\tau = (s_0, a_0, s_1, \dots, a_{T-1}, s_T)$ by applying a softmax over the per-action logits at each step on $\mathcal{A}(s_t)$. The forward policy $\pi_F$ is paired with a backward kernel $p_B$ taken as uniform over parents,
\begin{equation}
\log p_B(s_{t-1} \mid s_t) = -\log\big|\text{parents}(s_t)\big|,
\end{equation}
which is cheap to evaluate because the add-only grammar makes the parent set of any state directly enumerable. A state-conditional log-partition head $\log Z(s_0)$ is additionally trained, implemented as a two-layer MLP over the state encoding produced by the scorer. Conditioning $\log Z$ on $s_0$ allows a single sampler to handle the wide distribution of partial states produced by different briefs. The Trajectory Balance~\cite{malkin_trajectory_2023} residual for a trajectory $\tau$ is
\begin{equation}
\delta_{\text{TB}}(\tau) = \log Z(s_0) + \sum_{t=0}^{T-1} \log \pi_F(a_t \mid s_t) - \log R(s_T) - \sum_{t=1}^{T} \log p_B(s_{t-1} \mid s_t),
\end{equation}
and the GFlowNet loss is the mean squared residual over a batch of trajectories, $\mathcal{L}_{\text{TB}} = \mathbb{E}_{\tau}\!\big[\delta_{\text{TB}}(\tau)^{2}\big]$. Globally minimising $\mathcal{L}_{\text{TB}}$ aligns the forward marginal $\pi_F(s_T \mid s_0)$ with $R(s_T)$ up to the normaliser $Z(s_0)$.

\paragraph{Anchor and entropy regularisation.} Two auxiliary terms are introduced to stabilise the finetune. A \emph{KL anchor} to the frozen pretrain policy $\pi_{\text{pre}}$ prevents $\pi_F$ from drifting away from the corpus-grounded action distribution during reward chasing,
\begin{equation}
\mathcal{L}_{\text{KL}} = \mathbb{E}_{\tau, t}\!\big[\, \mathrm{KL}\big(\pi_F(\cdot \mid s_t)\,\|\,\pi_{\text{pre}}(\cdot \mid s_t)\big)\big],
\end{equation}
and an \emph{entropy bonus} on the per-state forward policy preserves exploration,
\begin{equation}
\mathcal{L}_{\text{H}} = \mathbb{E}_{\tau, t}\!\big[\, \mathbb{H}\big(\pi_F(\cdot \mid s_t)\big)\big].
\end{equation}
The total finetune loss is $\mathcal{L} = \mathcal{L}_{\text{TB}} + \lambda_{\text{KL}}\mathcal{L}_{\text{KL}} - \lambda_{\text{H}}\mathcal{L}_{\text{H}}$, where $\lambda_{\text{KL}}$ and $\lambda_{\text{H}}$ are small positive weights. To prevent $-\infty$ residuals from rare zero-reward trajectories, $\log R(s_T)$ is computed as $\log\max(R(s_T), \epsilon_R)$ with $\epsilon_R = 10^{-3}$.

\paragraph{Sampler architecture.} The scoring network $f_\theta$, shared across both stages, encodes a state by mean-pooling typed-node embeddings within each type and concatenating the pools with type-presence masks, where each node embedding concatenates a learned node-id embedding with a projected frozen CLIP-L text feature of its surface text. An action-scoring head maps the state encoding and a per-action feature vector to a scalar logit, and the $\log Z$ head is a two-layer MLP on the same encoding. The full architecture is in the supplement (Sec.~A.4).

\subsection{Reward Computation}
\label{sec:method:reward}
The terminal reward $R(s_T)$ aggregates three components evaluated on the rendered logo $\hat{x} = \mathcal{G}(\Theta(s_T))$ and on the symbolic state itself: a CLIP-based recognizability term, an aesthetics term judged by an MLLM, and a corpus-relative originality term that treats ``departure from existing logos'' as a first-class optimisation signal.
\paragraph{Recognizability and aesthetics.} Let $b_{\text{subj}}$ denote the subject phrase of the design brief. The recognizability term is the CLIP cosine similarity
\begin{equation}
  \mathrm{sim}(\hat{x}, b_{\text{subj}})
    = \cos\!\big(\phi_{\text{img}}(\hat{x}),\, \phi_{\text{txt}}(b_{\text{subj}})\big),
\end{equation}
The aesthetics term is computed as the normalised mean of three professional design subscores, namely \emph{balance}, \emph{colour}, and \emph{simplicity}, which are returned by a frozen multimodal LLM judge on a $[0, 10]$ scale, yielding $R_{\text{aes}}(s_T) = \tfrac{1}{30}(\text{balance} + \text{colour} + \text{simplicity}) \in [0, 1]$.
\paragraph{Corpus-relative originality.} We measure the extent to which $s_T$ departs from the corpus $\mathcal{D}$ along two complementary views: one operating on pixels and the other on symbols. The \emph{visual view} compares the rendered image against a precomputed cache of corpus-image embeddings,
\begin{equation}
R_{\text{vis}}(\hat{x}) = \mathrm{clip}\!\Big(1 - \max_{i \in \mathcal{D}}\cos\!\big(\phi_{\text{img}}(\hat{x}), \phi_{\text{img}}(x_i)\big),\, 0, 1\Big),
\end{equation}
which penalises rendered logos that closely resemble any existing logo in the corpus. The \emph{symbolic view} captures compositions whose rendered images may appear visually distant by chance, yet whose underlying pairings are commonplace: for each unordered pair $(u, v)$ of object-typed and visual-attribute nodes in $s_T$, we obtain the normalised co-occurrence frequency $\tilde{c}(u, v) \in [0, 1]$ from $G$ (Sec.~\ref{sec:method:grammar}) and average the pair-level surprise values,
\begin{equation}
R_{\text{sym}}(s_T) = \tfrac{1}{|\mathcal{P}(s_T)|}\sum_{(u, v) \in \mathcal{P}(s_T)} \big(1 - \tilde{c}(u, v)\big),
\end{equation}
where $\mathcal{P}(s_T)$ denotes the set of such pairs. The originality reward combines the two views with equal weights, namely $R_{\text{ori}}(s_T) = \tfrac{1}{2}R_{\text{vis}}(\hat{x}) + \tfrac{1}{2}R_{\text{sym}}(s_T)$.
\paragraph{Composition.} The three components are aggregated into a bounded weighted average,
\begin{equation}
R(s_T) = \frac{w_{\text{rec}}\, R_{\text{rec}} + w_{\text{aes}}\, R_{\text{aes}} + w_{\text{ori}}\, R_{\text{ori}}}{w_{\text{rec}} + w_{\text{aes}} + w_{\text{ori}}} \in [0, 1].
\end{equation}
\section{Experiments}
\label{sec:exp}

We evaluate DOGS against nine prompt-engineering baselines on two open-source T2I renderers, through two complementary batteries that measure logo quality and trademark infringement risk respectively. All methods are compared as prompt-side procedures. Each sends a user brief to a single renderer prompt and then to the same frozen models, so that every observed difference is attributable to the prompt rather than pixel synthesis.

\subsection{Experimental Setup}
\label{sec:exp:setup}

\paragraph{Training data.} The training corpus is drawn from L3D\cite{gutierrez-fandino_large_2021}. The $61{,}263$ re-annotated logos of Sec.~\ref{sec:method:space} are split into $58{,}243$ training and $3{,}020$ validation samples by a deterministic hash on the sample identifier (validation fraction $0.05$), and both training stages of Sec.~\ref{sec:method:sampler} inherit this split. For the visual originality term of Sec.~\ref{sec:method:reward}, every training logo is pre-rendered once through the evaluation renderer and its CLIP-image embedding is cached, yielding a fixed reference set of 61,263 vectors used to compute $R_{vis} $ during training.
\paragraph{Training procedure.} The sampler is trained in the two stages of Sec.~\ref{sec:method:sampler}, operating over the co-occurrence graph constructed as described in Sec.~\ref{sec:method:grammar}, which comprises $3{,}312$ nodes and $126{,}686$ edges and covers the majority of logo design scenarios. The supervised completion pretrain is optimised with Adam and stopped early by validation patience. The GFlowNet finetune follows a three-phase schedule: a warmup that trains $\log Z(s_0)$ alone, a main phase that linearly ramps the reward weight from $0$ to $1$ under a small KL anchor and entropy bonus, and a short anneal that lowers the softmax temperature and halves the regularisation weights. Full hyperparameters are in the supplement (Sec.~B).
\paragraph{Baselines.} We compare against nine alternatives, grouped by the two capabilities identified above. \emph{Original} passes the brief unchanged and supplies neither capability, serving as an anchor. Two \emph{LLM rewriters}, Qwen-3.5-9B and Llama-3.2-11B prompted as enhancers, supply both at once, as an instruction-tuned model can invent the missing design elements and phrase them in a single pass. Two \emph{trained optimizers}, Promptist~\cite{hao_optimizing_2023} and BeautifulPrompt~\cite{cao_beautifulprompt_2023}, refine and expand the surface phrasing of their input but introduce none of the design decisions the brief omits, and thus supply optimization only. The four \emph{LLM\,$\to$\,optimizer chains}, formed by piping a Qwen or Llama rewriter into Promptist or BeautifulPrompt, restore a division of labour in which the LLM completes the design and the optimizer refines it. We include them so that the optimizer-only methods are not penalised against completion-plus-optimization methods such as DOGS. 
\begin{figure}[t]
\centering
\footnotesize
\setlength{\tabcolsep}{4pt}
\renewcommand{\arraystretch}{1.0}

\begin{tabular}{@{}r@{\hspace{4pt}}cccc@{}}
\toprule
& \textbf{Original} & \textbf{Llama} & \textbf{Llama$\to$BP} & \textbf{DOGS (ours)} \\
\midrule
\rotatebox{90}{\parbox{1.7cm}{\centering\textbf{Comparison}}} &
\parbox[b]{0.18\linewidth}{\centering\scriptsize Logo of a paper, Red gold palette.\\[1pt]\includegraphics[width=\linewidth]{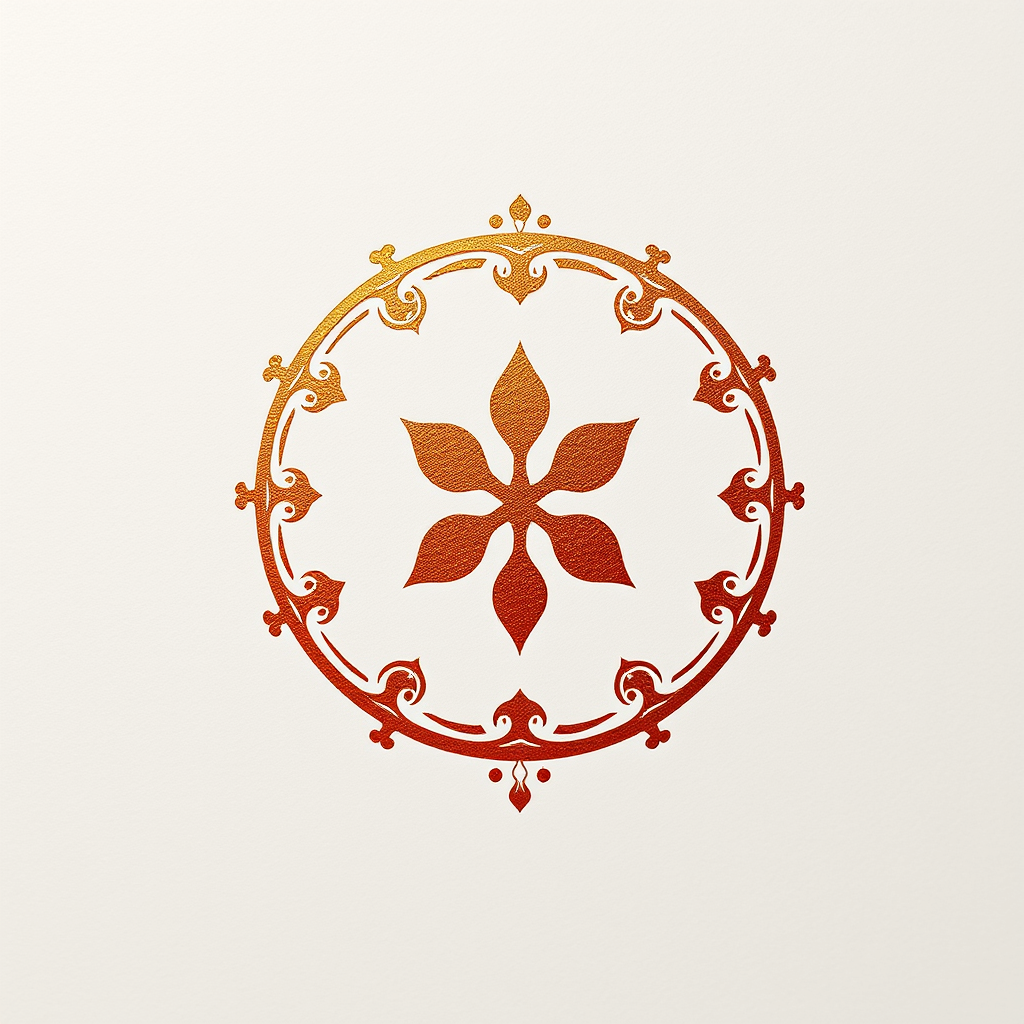}} &
\parbox[b]{0.18\linewidth}{\centering\scriptsize A stylized, modern logo of a paper scroll ...\\[1pt]\includegraphics[width=\linewidth]{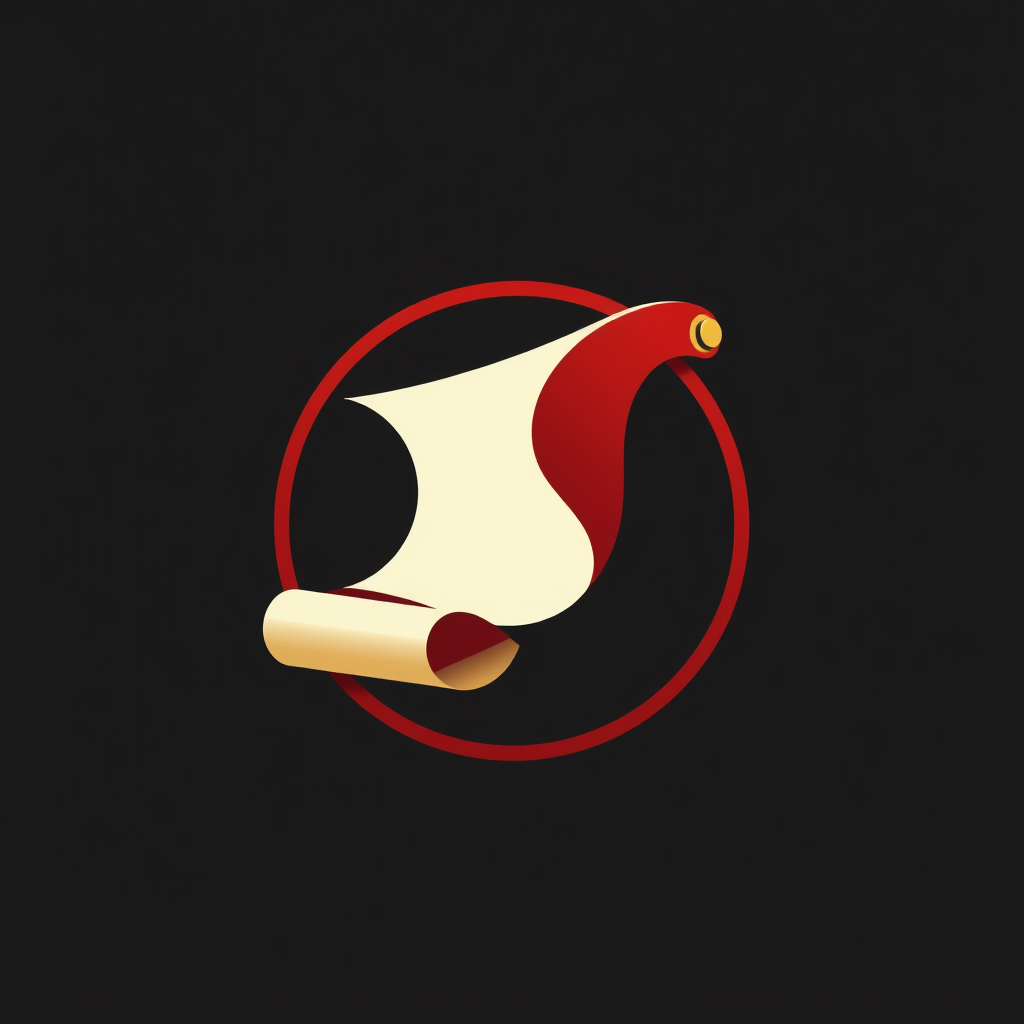}} &
\parbox[b]{0.18\linewidth}{\centering\scriptsize stylized modern logo of a paper scroll, ...\\[1pt]\includegraphics[width=\linewidth]{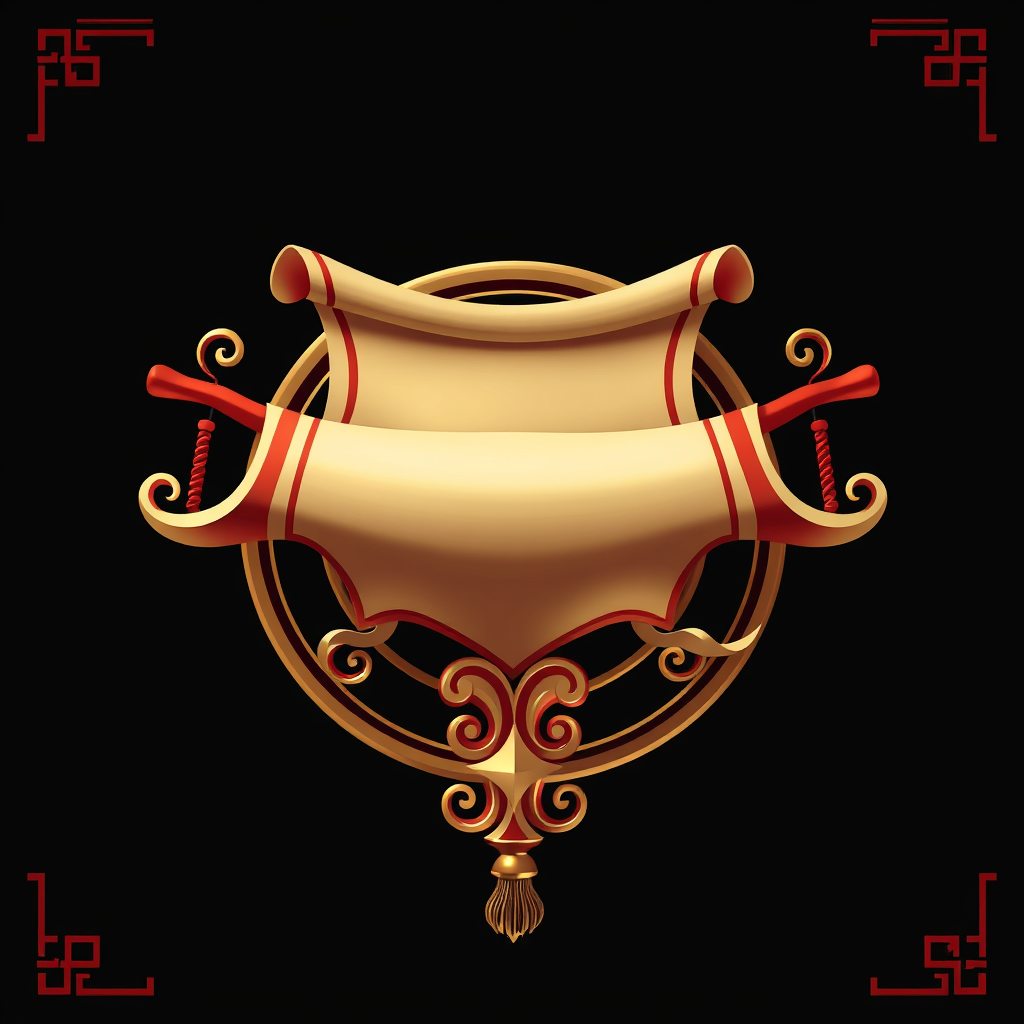}} &
\parbox[b]{0.18\linewidth}{\centering\scriptsize Logo of a paper inside a heart, an arrow ...\\[1pt]\includegraphics[width=\linewidth]{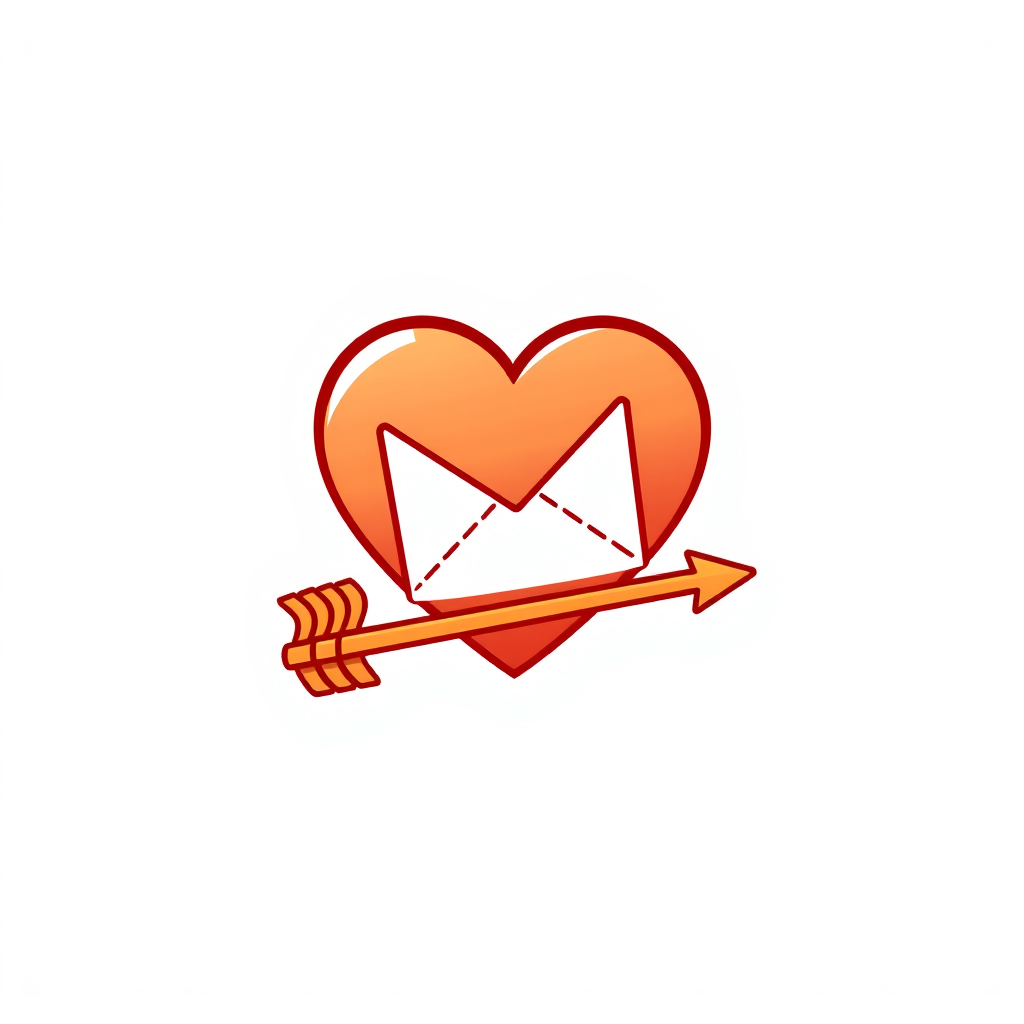}} \\[2pt]

\rotatebox{90}{\parbox{1.7cm}{\centering\textbf{Diversity}}} &
\includegraphics[width=0.18\linewidth]{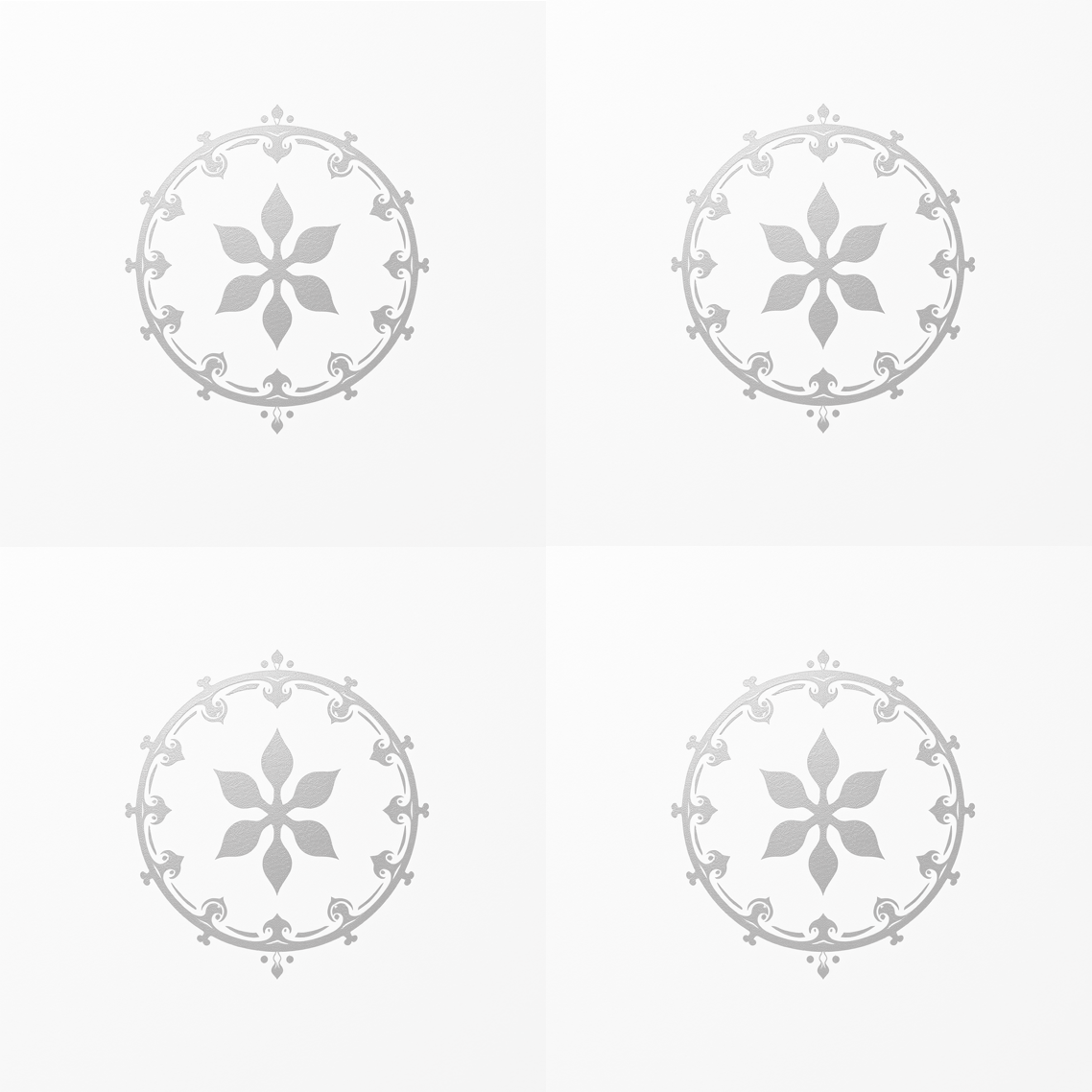} &
\includegraphics[width=0.18\linewidth]{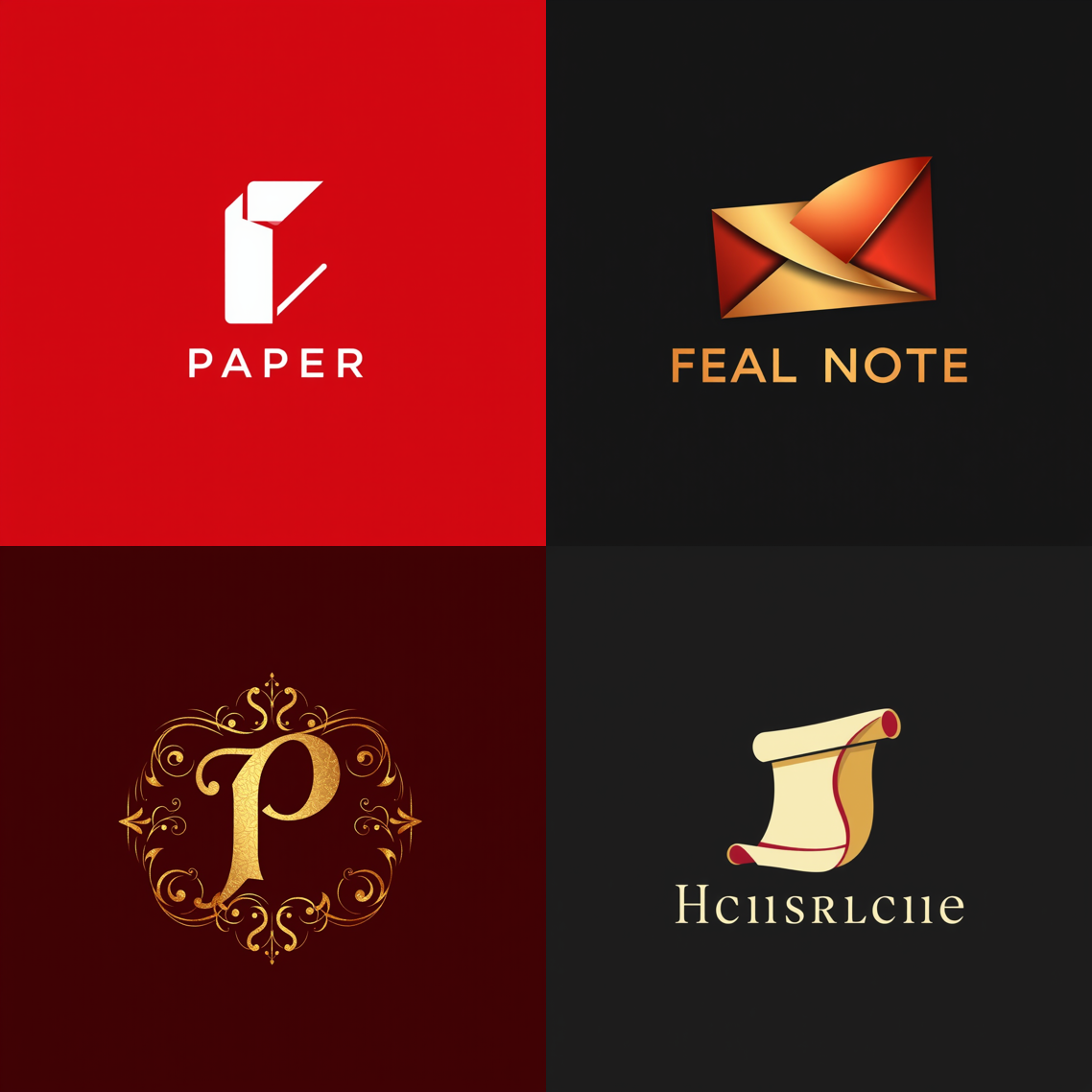} &
\includegraphics[width=0.18\linewidth]{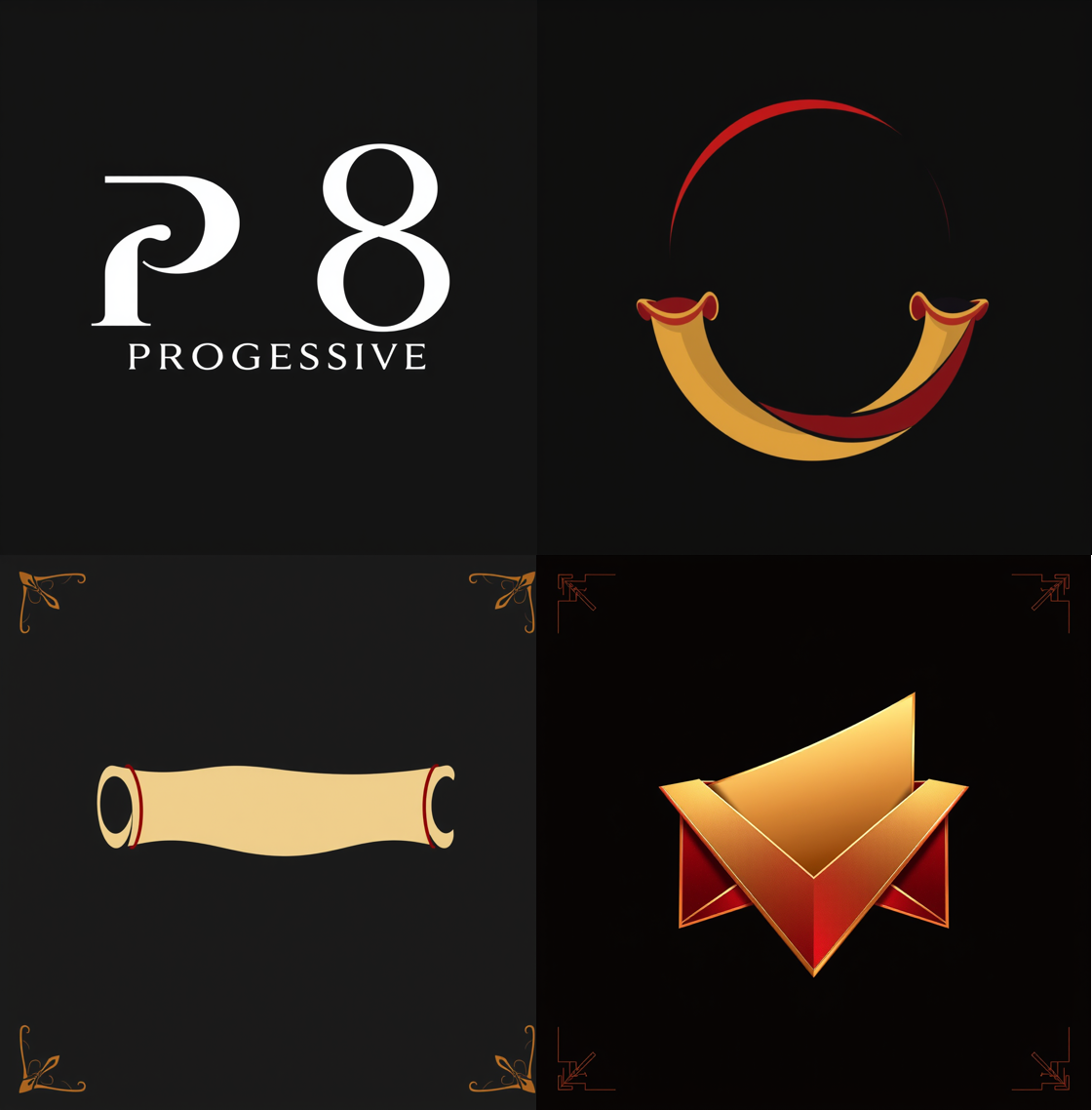} &
\includegraphics[width=0.18\linewidth]{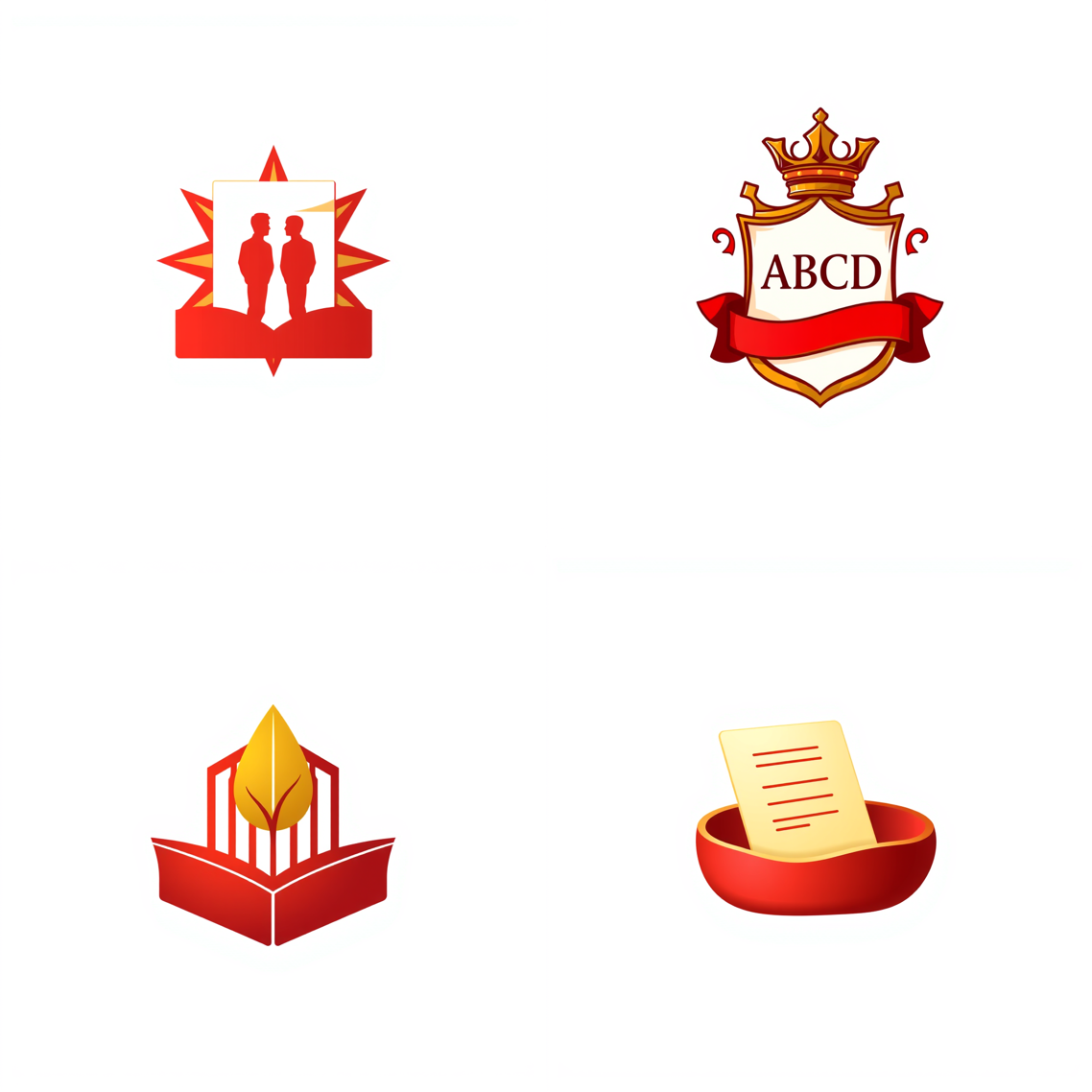} \\[2pt]

\rotatebox{90}{\parbox{1.7cm}{\centering\textbf{Infringement}}} &
\parbox[b]{0.18\linewidth}{\centering\scriptsize Mercedes-Benz logo\\[1pt]\includegraphics[width=\linewidth]{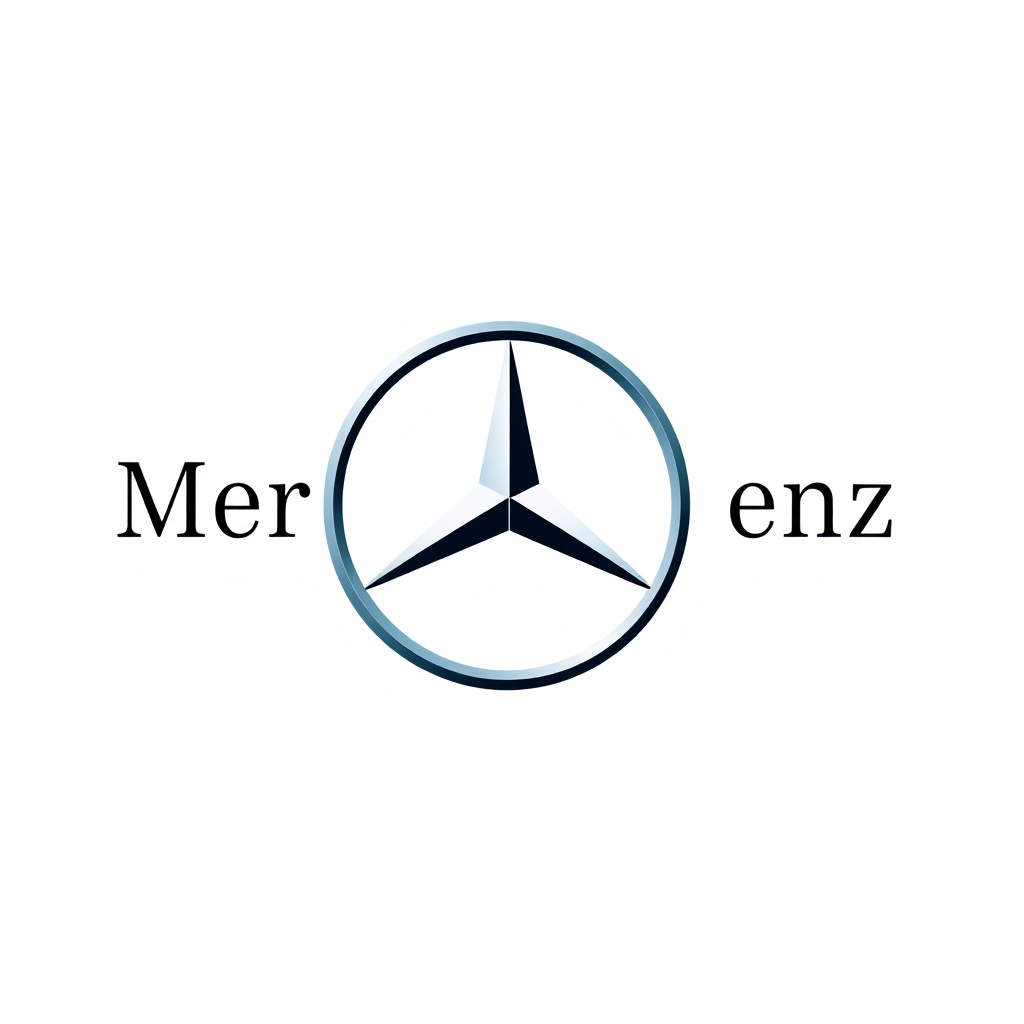}} &
\parbox[b]{0.18\linewidth}{\centering\scriptsize A sleek, modern Mercedes-Benz logo on ...\\[1pt]\includegraphics[width=\linewidth]{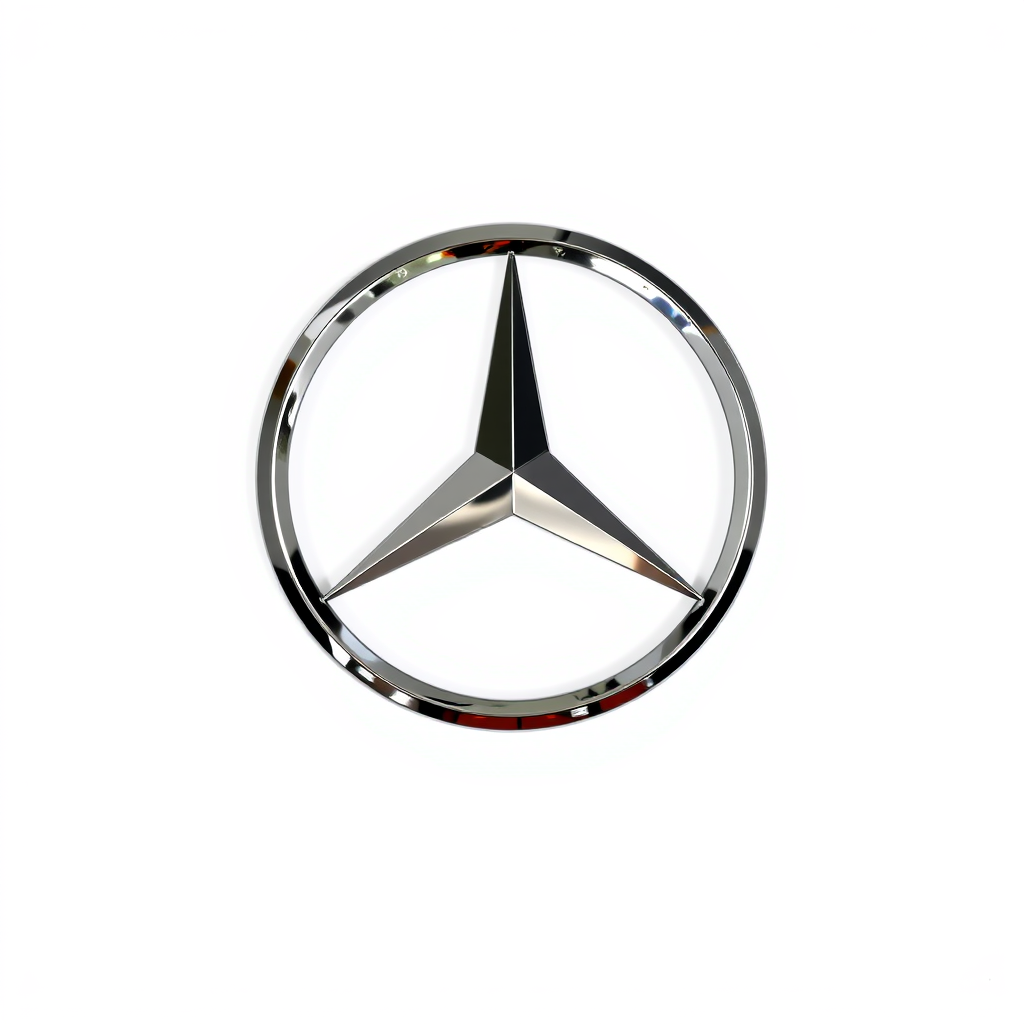}} &
\parbox[b]{0.18\linewidth}{\centering\scriptsize a stylish luxury Mercedes benz logo, ...\\[1pt]\includegraphics[width=\linewidth]{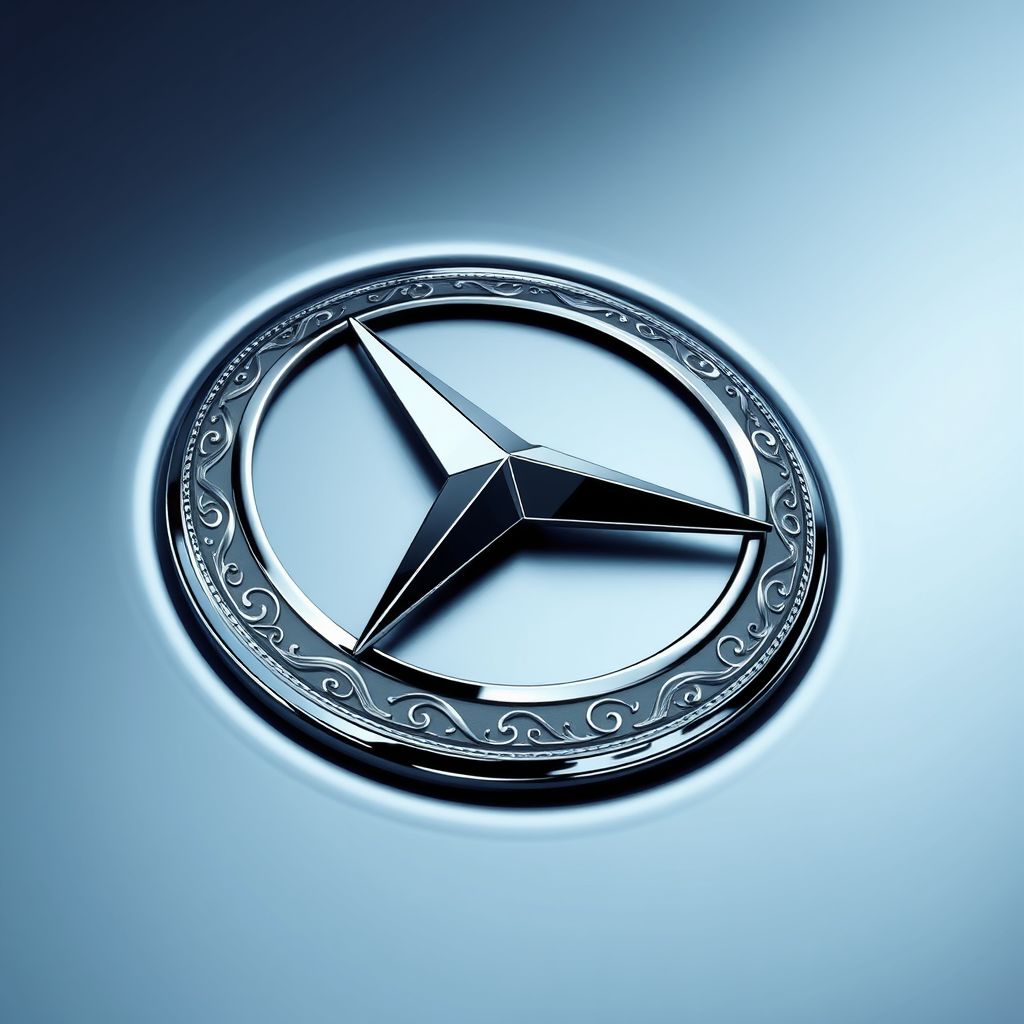}} &
\parbox[b]{0.18\linewidth}{\centering\scriptsize A logo featuring a shield, with a star ...\\[1pt]\includegraphics[width=\linewidth]{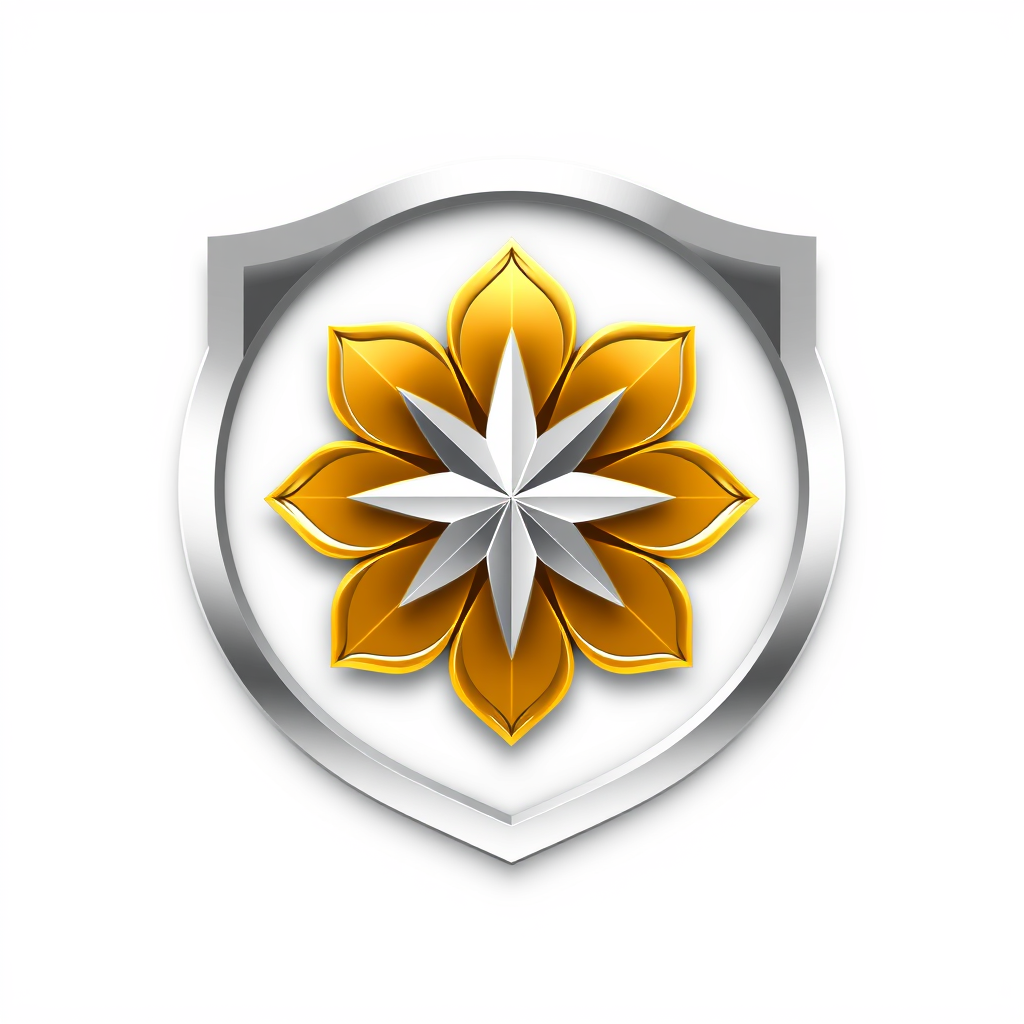}} \\
\bottomrule
\end{tabular}

\caption{\textbf{Qualitative comparison.} Columns are methods. \emph{Comparison}: DOGS composes a coherent design while baselines drift off-subject or render a stock image. \emph{Diversity}: four samples per method as $2{\times}2$ mosaics; DOGS yields four distinct yet usable logos, while baselines collapse or vary without comparable quality. \emph{Infringement}: a \emph{brand} prompt names a known trademark; baselines reproduce it, DOGS substitutes a different professional design (prompts truncated).}
\label{fig:qualitative}
\end{figure}
\paragraph{Renderers.} Each method is evaluated on two open-source T2I models, SDXL-Lightning (4-step, $768{\times}768$) and FLUX.1-schnell (4-step, $1024{\times}1024$). Because the two backbones differ in text encoder, latent geometry, and resolution, results are reported per renderer rather than averaged across renderers.
\paragraph{Evaluation data.} Since general-purpose prompt datasets contain no material dedicated to logo design, we build two held-out evaluation sets, both disjoint from the training corpus. The \emph{quality battery} uses $150$ natural-language logo briefs, generated by an LLM with reference to the $26$ categories of the Vienna Classification so as to span a broad range of subjects, with five samples drawn per cell, where a cell is a (prompt, method, renderer) combination. The \emph{infringement battery} uses a held-out $100$-brand reference set, each brand paired with its real logo and rendered under three prompt types that probe progressively weaker brand leakage: a \emph{brand} prompt names the trademark, a \emph{company} prompt describes the company without naming it, and a \emph{design} prompt supplies only visual cues, with one image generated per (brand, prompt-type, method, renderer) cell.

\subsection{Metrics}
\label{sec:exp:metrics}

\paragraph{Quality.} Each rendered logo is scored on three metrics, all rescaled to $[0, 10]$. Full formulae and calibration constants are in the supplement (Sec.~C).
\begin{itemize}\setlength{\itemsep}{1pt}
    \item \textbf{Recognizability ($Q_{\text{rec}}$).} The CLIP cosine between the rendered image and the text produced by the method.
    \item \textbf{Aesthetics ($Q_{\text{aes}}$).} The mean of two Qwen-3.5-27B subscores, \emph{overall} and \emph{simplicity}.
    \item \textbf{Generator diversity.} The mean pairwise CLIP-image distance among the five samples of a cell, computed with the renderer seed pinned across the five samples. Pinning the seed isolates the diversity contributed by the upstream method from the intrinsic noise of the renderer, so a deterministic method is correctly scored as zero.
\end{itemize}

\paragraph{Infringement.} This battery measures whether a rendered logo \emph{specifically} resembles an existing trademark rather than merely looking like a generic logo.
\begin{itemize}\setlength{\itemsep}{1pt}
    \item \textbf{Direct similarity.} The CLIP-image cosine between the generated image and the actual logo of the target brand.
    \item \textbf{Baseline similarity.} The same cosine to a randomly chosen other brand in the reference set, serving as a chance floor induced by shared structural priors.
    \item \textbf{Leakage lift $\Delta$.} Defined as $\Delta = \text{direct} - \text{baseline}$. A method that produces only a generic logo scores $\Delta \approx 0$, whereas a method that leaks the identity of the target brand scores $\Delta \gg 0$; subtracting the chance floor avoids rewarding methods that are merely stylistically generic.
\end{itemize}
Results are split by prompt type, since the three types probe different leakage regimes, from explicit invocation under a brand-name prompt to purely visual cues under a design-description prompt.

\subsection{Results}
\begin{table*}[t]
\centering
\footnotesize
\setlength{\tabcolsep}{4pt}
\renewcommand{\arraystretch}{0.95}
\caption{\textbf{Quality battery.} $150$ briefs, five samples per cell (metrics: Sec.~\ref{sec:exp:metrics}). Bold: best among baselines and DOGS; \emph{abl.} rows: ablations of Sec.~\ref{sec:exp:ablation}. Deterministic methods score zero diversity by construction.}
\begin{tabular}{l|ccc|ccc}
\toprule
& \multicolumn{3}{c|}{\textbf{SDXL-Lightning}} & \multicolumn{3}{c}{\textbf{FLUX.1-schnell}}\\
Method & $Q_{\text{rec}}\!\uparrow$ & $Q_{\text{aes}}\!\uparrow$ & Diversity\,$\uparrow$ & $Q_{\text{rec}}\!\uparrow$ & $Q_{\text{aes}}\!\uparrow$ & Diversity\,$\uparrow$ \\
\midrule
Original prompt          & 9.26\,{\tiny$\pm$0.15} & 4.03\,{\tiny$\pm$0.17} & 0.00\,{\tiny$\pm$0.00} & 9.70\,{\tiny$\pm$0.10} & 4.96\,{\tiny$\pm$0.17} & 0.00\,{\tiny$\pm$0.00} \\
\midrule
Qwen rewrite             & 7.41\,{\tiny$\pm$0.21} & 3.19\,{\tiny$\pm$0.15} & 2.25\,{\tiny$\pm$0.16} & 7.75\,{\tiny$\pm$0.20} & 3.64\,{\tiny$\pm$0.18} & 2.49\,{\tiny$\pm$0.16} \\
Llama rewrite            & 9.40\,{\tiny$\pm$0.11} & 4.07\,{\tiny$\pm$0.12} & 2.76\,{\tiny$\pm$0.18} & 9.71\,{\tiny$\pm$0.06} & 4.67\,{\tiny$\pm$0.11} & 2.30\,{\tiny$\pm$0.15} \\
Promptist                & 9.60\,{\tiny$\pm$0.09} & 3.35\,{\tiny$\pm$0.15} & 2.52\,{\tiny$\pm$0.20} & 9.50\,{\tiny$\pm$0.09} & 4.15\,{\tiny$\pm$0.15} & 2.33\,{\tiny$\pm$0.19} \\
BeautifulPrompt          & 9.61\,{\tiny$\pm$0.07} & 3.18\,{\tiny$\pm$0.11} & 3.34\,{\tiny$\pm$0.16} & 9.45\,{\tiny$\pm$0.08} & 3.65\,{\tiny$\pm$0.11} & 3.54\,{\tiny$\pm$0.16} \\
Qwen $\to$ Promptist     & \textbf{9.66}\,{\tiny$\pm$0.09} & 2.57\,{\tiny$\pm$0.15} & 2.26\,{\tiny$\pm$0.14} & 9.34\,{\tiny$\pm$0.12} & 3.10\,{\tiny$\pm$0.17} & 2.45\,{\tiny$\pm$0.17} \\
Qwen $\to$ BeautifulPrompt    & 9.63\,{\tiny$\pm$0.09} & 2.49\,{\tiny$\pm$0.13} & 2.08\,{\tiny$\pm$0.14} & 9.49\,{\tiny$\pm$0.13} & 3.05\,{\tiny$\pm$0.15} & 2.14\,{\tiny$\pm$0.15} \\
Llama $\to$ Promptist    & 9.47\,{\tiny$\pm$0.09} & 3.48\,{\tiny$\pm$0.12} & 3.23\,{\tiny$\pm$0.17} & 9.41\,{\tiny$\pm$0.08} & 4.32\,{\tiny$\pm$0.13} & 3.35\,{\tiny$\pm$0.18} \\
Llama $\to$ BeautifulPrompt   & 9.49\,{\tiny$\pm$0.08} & 3.13\,{\tiny$\pm$0.11} & 3.25\,{\tiny$\pm$0.15} & 9.38\,{\tiny$\pm$0.09} & 3.77\,{\tiny$\pm$0.13} & \textbf{3.60}\,{\tiny$\pm$0.16} \\
\midrule
Graph retrieval (abl.)   & 9.31\,{\tiny$\pm$0.15} & 4.01\,{\tiny$\pm$0.16} & 0.00\,{\tiny$\pm$0.00} & 9.65\,{\tiny$\pm$0.13} & 4.43\,{\tiny$\pm$0.16} & 0.00\,{\tiny$\pm$0.00} \\
DOGS (pretrain, abl.)    & 9.38\,{\tiny$\pm$0.10} & 4.27\,{\tiny$\pm$0.10} & 3.05\,{\tiny$\pm$0.20} & 9.74\,{\tiny$\pm$0.07} & 4.59\,{\tiny$\pm$0.09} & 2.79\,{\tiny$\pm$0.19} \\
\underline{DOGS (ours)}     & \underline{9.40\,{\tiny$\pm$0.09}} & \underline{\textbf{4.09}\,{\tiny$\pm$0.08}} & \underline{\textbf{3.71}\,{\tiny$\pm$0.18}} & \underline{\textbf{9.79}\,{\tiny$\pm$0.05}} & \underline{\textbf{4.81}\,{\tiny$\pm$0.08}} & \underline{3.41\,{\tiny$\pm$0.18}} \\
\bottomrule
\end{tabular}
\label{tab:quality}
\end{table*}

\paragraph{Quality.} Table~\ref{tab:quality} reports the quality battery, which asks whether a prompt-side method can supply the three properties a usable logo demands at once: faithful recognizability of the subject, professional aesthetics, and diversity across renderings of one brief. Recognizability is the primary requirement, and the CLIP-cosine scoring compresses the already-high subject--image agreement into a narrow band, which showcases the renderer's capacity. DOGS is the only completion method whose recognizability rises sharply from the weaker SDXL to the stronger FLUX ($9.40 \to 9.79$, the top FLUX score), while the optimizer family moves the other way (Qwen\,$\to$\,Promptist $9.66 \to 9.34$). DOGS's completed designs are thus faithfully recognizable once the renderer can execute them. The 4-step $768^2$ SDXL realises that detail only partially, explaining its mid-band score.

On aesthetics, DOGS matches or exceeds the LLM rewriters on both renderers ($4.09$ vs Llama $4.07$ on SDXL; $4.81$ vs $4.67$ on FLUX), indicating that a general LLM without professional design priors cannot substitute for a domain-specific design model. Its margin over the trained optimizers is far larger (Promptist and BeautifulPrompt fall to $3.18$--$3.35$ on SDXL and the LLM\,$\to$\,optimizer chains collapse to $2.49$--$2.57$), which shows that optimisation against generic image-quality steers phrasing orthogonally, even detrimentally, to logo aesthetics. \emph{Original} reaches the highest FLUX aesthetics ($4.96$), but it is deterministic (diversity $0$) and performs no completion, so its score reflects image-level scorers favouring a single, simple output rather than a solution to the design task. Once completion is added in, DOGS is the only method that leads on aesthetics and diversity jointly. Every baseline lacks in at least one of the three properties.

\paragraph{Infringement.} Table~\ref{tab:infringement} reports the infringement battery, which probes whether a method reproduces a \emph{specific} existing trademark by sweeping three prompt types that invoke the brand with decreasing explicitness, so as to locate where each method begins to leak. Leakage is largest under the \emph{brand} prompt, where the trademark name enters the input directly. Every baseline spikes there (\emph{Original} $0.13$ / $0.20$) and even the strongest chain stays at $0.05$ / $0.09$, whereas for DOGS $\Delta$ stays at $0.02$ on both renderers which is a five- to tenfold reduction over the raw brief. This is due to the design of DOGS (Sec.~\ref{sec:method:sampler}): the design-space whitelist admits no trademarked names so brand cues are dropped at parse time. Plus, the originality reward further penalises proximity to corpus logos.

As the prompt weakens from \emph{company} to \emph{design}, baseline leakage falls toward the floor and the gap narrows, yet DOGS stays flat at $\approx 0.02$ across all three regimes on both renderers, so its non-infringement does not depend on how much brand signal the prompt carries. The one column it does not lead is SDXL\,/\,\emph{design} ($0.03$ vs $0.01$). A pure-design prompt asks for the same cues that the reference carries. So, as a function of design fidelity, rendering them faithfully brings the output marginally closer to the original while the gap stays far below any infringement threshold.

\begin{table*}[t]
\centering
\footnotesize
\setlength{\tabcolsep}{4pt}
\renewcommand{\arraystretch}{0.95}
\caption{\textbf{Infringement battery.} Leakage lift $\Delta = \text{direct} - \text{baseline}$, $100$ brands, one image per cell. Columns are the three prompt types of Sec.~\ref{sec:exp:setup}.}
\begin{tabular}{l|ccc|ccc}
\toprule
& \multicolumn{3}{c|}{\textbf{SDXL-Lightning}} & \multicolumn{3}{c}{\textbf{FLUX.1-schnell}}\\
Method & Brand\,$\downarrow$ & Company\,$\downarrow$ & Design\,$\downarrow$ & Brand\,$\downarrow$ & Company\,$\downarrow$ & Design\,$\downarrow$ \\
\midrule
Original prompt          & 0.13\,{\tiny$\pm$0.02} & 0.04\,{\tiny$\pm$0.01} & 0.05\,{\tiny$\pm$0.01} & 0.20\,{\tiny$\pm$0.02} & 0.08\,{\tiny$\pm$0.01} & 0.06\,{\tiny$\pm$0.02} \\
\midrule
Qwen rewrite             & 0.07\,{\tiny$\pm$0.02} & 0.02\,{\tiny$\pm$0.01} & 0.02\,{\tiny$\pm$0.01} & 0.11\,{\tiny$\pm$0.02} & 0.04\,{\tiny$\pm$0.02} & 0.03\,{\tiny$\pm$0.02} \\
Llama rewrite            & 0.12\,{\tiny$\pm$0.01} & 0.04\,{\tiny$\pm$0.01} & 0.05\,{\tiny$\pm$0.01} & 0.18\,{\tiny$\pm$0.02} & 0.07\,{\tiny$\pm$0.01} & 0.07\,{\tiny$\pm$0.02} \\
Promptist                & 0.10\,{\tiny$\pm$0.02} & 0.04\,{\tiny$\pm$0.01} & 0.04\,{\tiny$\pm$0.01} & 0.19\,{\tiny$\pm$0.02} & 0.07\,{\tiny$\pm$0.02} & 0.07\,{\tiny$\pm$0.02} \\
BeautifulPrompt          & 0.10\,{\tiny$\pm$0.02} & 0.04\,{\tiny$\pm$0.01} & 0.02\,{\tiny$\pm$0.01} & 0.15\,{\tiny$\pm$0.02} & 0.07\,{\tiny$\pm$0.01} & 0.03\,{\tiny$\pm$0.01} \\
Qwen $\to$ Promptist     & 0.06\,{\tiny$\pm$0.02} & 0.02\,{\tiny$\pm$0.01} & 0.02\,{\tiny$\pm$0.01} & 0.11\,{\tiny$\pm$0.03} & 0.06\,{\tiny$\pm$0.02} & 0.04\,{\tiny$\pm$0.02} \\
Qwen $\to$ BeautifulPrompt    & 0.05\,{\tiny$\pm$0.02} & 0.02\,{\tiny$\pm$0.01} & \textbf{0.01}\,{\tiny$\pm$0.01} & 0.09\,{\tiny$\pm$0.02} & 0.04\,{\tiny$\pm$0.01} & 0.02\,{\tiny$\pm$0.01} \\
Llama $\to$ Promptist    & 0.08\,{\tiny$\pm$0.02} & 0.04\,{\tiny$\pm$0.01} & 0.03\,{\tiny$\pm$0.01} & 0.16\,{\tiny$\pm$0.02} & 0.07\,{\tiny$\pm$0.02} & 0.04\,{\tiny$\pm$0.01} \\
Llama $\to$ BeautifulPrompt   & 0.08\,{\tiny$\pm$0.02} & 0.03\,{\tiny$\pm$0.01} & 0.02\,{\tiny$\pm$0.01} & 0.12\,{\tiny$\pm$0.02} & 0.05\,{\tiny$\pm$0.01} & 0.04\,{\tiny$\pm$0.01} \\
\midrule
\underline{DOGS (ours)}     & \underline{\textbf{0.02}\,{\tiny$\pm$0.01}} & \underline{\textbf{0.02}\,{\tiny$\pm$0.01}} & \underline{0.03\,{\tiny$\pm$0.01}} & \underline{\textbf{0.02}\,{\tiny$\pm$0.01}} & \underline{\textbf{0.02}\,{\tiny$\pm$0.01}} & \underline{\textbf{0.02}\,{\tiny$\pm$0.01}} \\
\bottomrule
\end{tabular}
\label{tab:infringement}
\end{table*}

\subsection{Ablation study}
\label{sec:exp:ablation}
 
We isolate the contribution of each learning stage with two ablated variants, reported as the \emph{abl.} rows of Table~\ref{tab:quality}. \emph{Graph retrieval} removes all learning: given a partial state it returns the caption of the structurally nearest training logo, keeping the design space but replacing the sampler with a deterministic lookup. \emph{DOGS (pretrain)} keeps the design space and the supervised completion policy but drops the reward-conditioned GFlowNet finetune of Sec.~\ref{sec:method:sampler}.
 
Graph retrieval already attains corpus-level recognizability and aesthetics ($9.31$/$4.01$ on SDXL, $9.65$/$4.43$ on FLUX), confirming that the design space alone yields sound logos. But, it scores zero diversity since this method is deterministic. Adding the supervised sampler recovers diversity ($3.05$/$2.79$) and improves quality close to its final level. So, design completion through a learned policy supplies most of the quality gain. The reward-conditioned finetune then drives diversity to its highest values ($3.71$/$3.41$) and yields the best recognizability on both renderers, while aesthetics is essentially preserved, improving on FLUX ($4.59 \to 4.81$) and trading a small margin on SDXL ($4.27 \to 4.09$) for the substantial diversity gain. Each stage thus contributes a distinct effect, with reward-proportional sampling specifically responsible for diversity.

\section{Conclusion}
\label{sec:conclusion}
We recast prompt optimization for logo design from language-space rewriting to sampling within a structured design space, realized as DOGS: a corpus-induced graph grammar over typed design elements, completed by a GFlowNet sampler that draws design graphs in proportion to a reward over recognizability, aesthetics, and corpus-relative originality. Thereby, DOGS has infringement avoidance built into the sampler, obviating the need for post-hoc filters. Across two renderers and nine baselines, DOGS yields logos that are more recognizable, aesthetic, and diverse, and far less prone to infringement, suggesting the design-space formulation extends naturally beyond logos to structured visual design.


\bibliography{egbib}

\end{document}